\documentclass[runningheads]{llncs}
\usepackage[T1]{fontenc}
\usepackage{graphicx}
\begin{document}
\title{Challenges in Evaluating Explanation Methods for Static and Evolving Data} 
\titlerunning{Challenges in evaluating explanation methods ...}

\author{Jerzy Stefanowski\orcidID{0000-0002-4949-8271} }
\authorrunning{Jerzy Stefanowski}
\institute{Poznan University of Technology, Institute of Computing Sciences, Poland
\email{jerzy.stefanowski@cs.put.poznan.pl} }

\maketitle

\begin{abstract}
This paper  addresses the limitations of Explainable Artificial Intelligence (XAI) with respect to insufficient evaluation. They are illustrated through the DetoxAI image recognition system for bias detection and concept unlearning. Then, an example of a human-grounded evaluation of methods for explaining image classification is presented. The paper further explores methods for adapting explanations to evolving data streams with concept drift. Experiences with adapting  counterfactuals for this problem are discussed.  Finally it is related to the challenges of tracking the co-evolution of data, models, and explanations.\footnote{This paper has been accepted for a publication in J.Nalepa (ed) Explainable AI in Space. Proceedings of EASi 2026 Workshop at IJCAI-ECAI 2026 Bremen, Springer  CCIS vol 3107 (2016).}

\end{abstract}
%
%

%\maketitle

%
\section{Introduction}

The development of Artificial Intelligence systems in recent decades has led to many successes, but also to a growing awareness of the risks and the need to meet additional requirements.  Many of them concern Responsible or Trustworthy AI, where the need for transparency in the systems and the ability to explain how they work become important \cite{kaur2022trustworthy}. The current Machine Learning (ML)systems are often based on more and more complex deep networks, which are opaque to users and do not provide such explanations. 

This lack of explainability poses an obstacle to effective application in many fields, especially where high-stakes decisions involving people are made. In particular, it concerns medical applications, at least for the following reasons \cite{Sadehavi2024}: 
\begin{itemize}
\item Explainability improves clinical trust: physicians must verify the clinical logic behind an output before  high-risk treatments.
\item It may discover some errors, dataset biases and false correlations, preventing wrong automatic diagnostic decisions.
\item It supports legal accountability: medical responsibility lies with humans; so physician should  require interpretable evidence to justify their actions.
\item It could be used for scientific discovery: explainable models can highlight novel patterns, advancing medical research and education. % of students or younger physicians.
\end{itemize}

Similarly, explanations are extremely important for critical systems (including aerospace) due to mission-critical verification and failure forensics, extreme domain adaptation or strict regulatory compliance. As  many such systems are autonomous and are deployed in high-stakes environments, the 'black-box' nature of deep learning introduces severe operational risks. For instance, in mission-critical aerospace applications, such as spacecraft health monitoring and anomaly detection, integrating explainability is a strict prerequisite to guarantee operational trust and system usability \cite{Tahir2025}. 

The above arguments provide motivations for the development and use of Explainable Artificial Intelligence (XAI) \cite{guidotti2018survey}.  Unfortunately, in XAI there is now a trend to constantly introduce more and more new methods, while the assessment of their usefulness is too limited \cite{moshkovitz2026explainability}. Moreover, the number of comparative studies of XAI methods is too limited and the spectrum of studied measures is too narrow. In a certain sense, this resembles the critical situation from the beginning of the year surrounding the development of supervised classification methods and their actual usefulness, which, to paraphrase David Hand’s famous text, is more of an “illusion of progress” \cite{hand2006classifier} 

In our opinion more research attention should be devoted to experimental evaluations, both quantitative (measures for more automatic studies) and qualitative (human or application grounded) ones and to the appropriate interaction with human experts. This talk will firstly review the current proposals for evaluating local explanations, and by case studies will discuss how to evaluate them better also with designing appropriate human surveys. 

The other topic discussed in this talk includes using XAI for explaining changes in data streams, where data distributions and models evolve over time – which is commonly studied as concept drift \cite{Lu2018}. Recall that most of standard XAI methods are designed for static settings, where models are trained once and explanations are generated for fixed data and model \cite{hinder2023model}. Applying such static XAI to evolving data leads to failures, as explanations may become stale, inconsistent over time and misleading.  This talk will present the speaker's latest original works and experiences with adapting prototypes and counterfactuals for this context, and to better characterize causes of the concept drift. Finally, it is related to challenges of evaluation of such aspects. 

\section{Difficulties in Evaluating XAI Methods}
\label{sec2:methods}

Evaluating the explanations is challenging, given the wide range of numerous proposals and ill-defined tasks of XAI \cite{moshkovitz2026explainability}.  

Firstly,  the XAI literature has produced a wide range of mainly post-hoc techniques to approximate different aspects of model performance with many methodological paradigms. However, they differ very much in terms of the types of data (tables, images, text, time series, etc.), the representations of explanations provided, its focus on different audiences, and other aspects.  As discussed in the paper \cite{moshkovitz2026explainability}, despite substantial methodological diversity, XAI approaches are not easy to apply, are highly sensitive to arbitrary design choices, and are often misused in practice. In general, they are  used correctly by model developers or domain experts often than by other users. 

The concept of XAI is rather ill-defined; see the overview of various definitions in \cite{arrieta2020explainable}. Here, we follow the definition  \cite{guidotti2018survey}: \textit{Explainable-AI explores and investigates methods to produce or complement AI models to make accessible and interpretable the internal logic and the outcome of the algorithms, making such process understandable by humans}. 
So, we can distinguish between \textit{global explanations} and \textit{local explanations} --  which we will further understand as attempts to identify the reasons behind a black-box ML model’s decision for a specific instance. Further on, we will consider local explanations.

Another difficult aspect is that explanations are intended for people and should be useful to them. This is also  ambiguous, as it depends strongly on the type of recipient, including their domain knowledge, preferences, and the specific task being solved. Following \cite{byrne2019counterfactuals}, currently proposed XAI methods take too much into account the perspective of the AI system developers, not other users. 

Furthermore, human evaluation is underutilized in assessing the effectiveness of these proposals. The review paper \cite{Lopes2022} cites research on recently published articles, according to which 97\% of the reviewed works pointed out that such explanations serve users, but most of them did not evaluate them through any user study. Similarly, other authors reviewed 381 XAI papers and found that only 5\% explicitly focused on a deeper evaluation of XAI methods. The paper \cite{Lopes2022} also point out other difficult issues such as: Lack of scientific consensus on evaluations;  Lack of multi-disciplinarity (needs for integrating HCI, Cognitive Science, Psychology); Lack of standarized procedures and  Lack of good tools for visualization and human interaction

Current taxonomies on evaluation of XAI generally distinguish between \textit{Human - centered} and \textit{Computer - centered} ones. They are influenced by Doshi-Velez and Kim’s proposal \cite{doshivelez2017rigorousscienceinterpretablemachine}, which considered participation of humans in XAI evaluation. This taxonomy is structured as follows:
\begin{itemize}
\item \textit{Application-grounded evaluation} (real tasks) - requires that human experts conduct tasks and experiments in a real setting; it can verify how well explanations assist humans trying to complete real application tasks.
\item \textit{Human-grounded evaluation} (simpler tasks, lay user) - Humans (rather laymen  being not experts) conduct simpler subject experiments with simplified tasks. It could cover a bigger subject pool, often human surveys.
\item \textit{Functionally-grounded evaluation} (proxy tasks) - requires no human experiments. Selected measures are used as formal proxy to evaluate some properties of  explanations.
This is a typically used automatic experimental approach (currently dominating in XAI literature).
\end{itemize} 

Considering Human Centered Evaluation, the main postulated categories could cover: User trust, understandability, explanation usefulness and satisfaction, cf. \cite{Lopes2022}. 

The desiderata for the automatic evaluation often concern comparing the accuracy of surrogate (explanation) model $f$ with the original black box model $h$. Usually they are realized by proxy measures assessing the following issues:
\begin{itemize}
\item Fidelity – to what extent are the predictions of $f$ consistent with model $h$.
\item Identity – identical examples should have identical explanations.
\item Stability – the same explanations for the similar examples $x$.
\item Importance of features for predictions.
\item Compactness and simplicity of the explanation representation.
\item Representativeness – does explanation covers enough learning examples.
\end{itemize}

Furthermore, for each category of explanations (e.g., image salience maps, counterfactuals, feature attributes) a set of specific properties and measures is used; see, e.g, counterfactuals Section  \ref{sec:counter}.

Although some of the above measures are often used in the XAI literature, fulfilling them does not always translate into the usefulness or correctness of the explanations themselves. In his paper \cite{Guidotti2021}, Guidotti experimentally demonstrated that focusing on high fidelity when evaluating proxy explanations may be incorrect with respect to the background knowledge, for instance rule-based explainers failed in retrieving all the features in the rule conditions when the number of features increases; Features importance: explainers failed in considering all the relevant features and in assigning the correct weight to them.

The above issues are also related to the insufficient number of well-prepared benchmark datasets with precise annotations of the expected explanations. Furthermore, works such as \cite{Guidotti2021,stefanowski2023} propose the use of multiple evaluation metrics in the automatic evaluation of XAI methods.

\section{Two Cases: How to Better Evaluate}

We will briefly present two case studies of challenges in evaluating explanation methods for image recognition deep neural networks, which differ from one another. The first case study demonstrates the use of XAI in the context of automated methods, while the second illustrates the challenges of a human-grounded evaluation study.

\subsection{DetoxAI: Using XAI to Improve Fairness}

\begin{figure}[tb]
    \centering
    \includegraphics[width=0.95\textwidth]{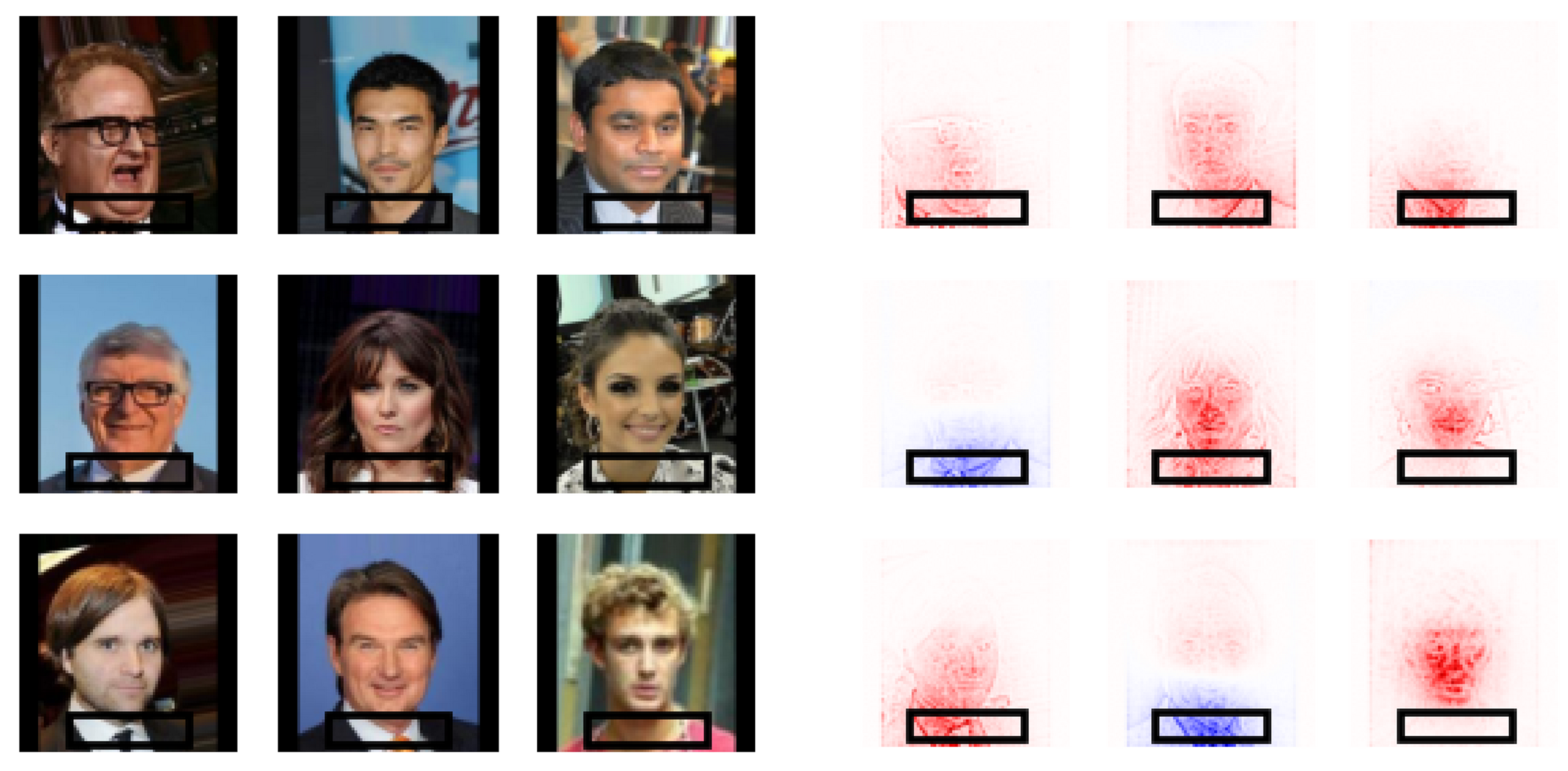}
    \caption{The left panel shows raw face images, and the right panel, corresponding CRP saliency maps. In the saliency maps, red hues indicate positive relevance the true class, while blue hues indicate negative contributions - more details in \cite{sztukiewicz2025investigating}.}
    \label{fig:enter-label}
\end{figure}

The first case shows using XAI  to detect bias in the data and in the performance of a convolutional neural network (CNN), followed by actions aimed at effectively improving the system’s performance, the effectiveness of which can be measured using selected metrics.

It  is devoted to  fairness and bias in decisions of ML systems.  This becomes particularly problematic when dealing with so called protected attributes (such as gender or race), which are high-level features (concepts) that are inaccessible to ML models operating only on pixel data without explicit features creation. In particular image neural networks may develop internal representations that encode not only useful high-level features but also harmful biases \cite{bolukbasi_2016}.

In  \cite{sztukiewicz2025investigating} we have considered several different models of CNN (different Resnet architectures) and human face datasets and observed predictions with respect to several measures, including  group fairness. In in the CelebA dataset, we discovered that wearing some items, such as a necktie, is highly correlated with the person gender. It means that the 'wearing necktie feature' can be used as a proxy feature to infer gender, thus creating potential unintended pathways for discrimination. It is illustrated in Fig. \ref{fig:enter-label}. Moreover, it has observed that the models consistently misclassifies individuals wearing neckties as not smiling, revealing unintended bias and decreasing values of selected fairness measures.

The first challenge was how to detect this bias. In the DetoXAI system \cite{stkepka2025detoxai}, we used XAI methods based on Saliency Maps (which highlight regions of the input image that influence the network’s predictions) -- see Fig. \ref{fig:enter-label}. We considered various methods for generating them, finally directed to using Concept-wise Relevance Propagation (CRP) \cite{CRP2023}, where relevance scores are propagated backward through the network layers, while also allowing us to identify which neurons (based on their scores) are most closely related to the undesired concept (bias). Its quantitative evaluation can be performed more automatically using new metrics for saliency maps introduced in \cite{sztukiewicz2025investigating}.

Another challenge is deciding how to eliminate this unwanted bias. Since modifying the training dataset and retraining the network from scratch is costly, we decided to use concept unlearning methods. We tested several recent post-hoc methods, such as those based on Savani and Zhang, LEACE, AClarc, and post-hoc threshold optimization \cite{mehrabi_2021_survey}. 

The results of the experiments described in \cite{sztukiewicz2025investigating} clearly showed that, although the methods differed, most of them led to the elimination of the network’s undesirable focus on biases—which can be observed qualitatively by a visual inspection of saliency maps and quantitatively in improvement of selected group-level fairness metrics (such as EqualizedOdds and Demographic Parity) for the network’s predictions. 

\subsection{Human Study on Evaluating Image Recognition XAI} 

The next case involves a discussion of the challenges of conducting human-grounded evaluation -- see also the previous section \ref{sec2:methods}. Here, in line with this taxonomy, we consider non-expert humans performing simpler tasks.  It concerns the evaluation of various XAI methods used to explain the predictions of convolutional neural networks—which are CNNs (ResNet). This case stems from a project carried out as a master’s thesis by Ms. Zofia Frackowiak \cite{Frackowiak2023}.

Given the available implementations, we considered three different XAI methods (each based on different principles, where the first two focus on searching for concepts in the data) that generate salience maps on the processed images:
\begin{itemize}
\item ACE - Automatic Concept-based Explanations with Concept Activation Vectors
\item ProtoPNet - Prototypical Network
\item RISE - Randomized Input Sampling for Explanations
\end{itemize}

We decided to develop an online survey targeting students and research staff from various universities with varying levels of experience in the use of AI, image recognition, and XAI methods.

The first challenge was deciding which task and the image dataset that would be understandable to a broad group of respondents. We first analyzed several popular benchmark datasets and specific subcategories of objects. An important lesson learned from conducting a preliminary pilot study with participants, led us to focus on animals and to select 10 very well-known, easily recognizable, and diverse animals from the ImageNet (such as an elephant, crocodile, hippopotamus, panda, tiger, zebra, lion, bison, frog, and others). Discussions with the pilot study participants indicated that these are animals that are relatively easy to recognize; the participants were also able to assess the presence of specific anatomical features.

Using a large subset of these animals (many versions for the same animal) as training data, the CNN was trained, achieving high recognition accuracy. For the final survey we selected a few different kinds of photos for each animal. 

Another challenge was to standardize the visualizations provided by the implementations of the selected methods. It was discovered that, although all of them returned salience maps, their graphical formats differed radically -- which, during the initial pilot tests, led to confusion among respondents and made it more difficult to compare results of implemented methods. Thus, Ms Frackowiak converted the results from the original salience maps into a uniform graphical interface (using a colored line and more distinct highlighting image). %This required certain design decisions, such as omitting less frequent pixels of other elements in the image, and fine-tuning each method to ensure they attempt to identify more consistent regions. 

In the final survey, for selected  photos of specific animals, up to six explanations derived from different runs of the explanation methods were shown—without informing the respondent about the method’s name and asking them to select 0 to 2 of the best ones (which were later assigned ranking points).

Moreover, it was decided to collect respondent’s basic demographic information (age, education, familiarity with AI and ML methods, familiarity with some XAI methods, and experience taking good photos) + special questionnaires to determine which of the anatomical animal features should be considered as most important for recognizing a given animal in the presented image.

A total of 148 completed surveys were collected from a very diverse audience and then analyzed. The main result (expressed as a number of ranking points) was the respondents’ preference for the ProtoPNet method (1,045 total points), followed by ACE (763) and RISE (643) overall for all presented animals. With regard to images of individual animals, the ProtoPNet method won in most cases, but the point differences depended on the specific animal (the strongest victory was for the zebra, panda, and tiger), while the ACE method won for the bison and was nearly a tie for the elephant.
An additional analysis of different demographic responses did not reveal significant variation in the distribution of responses across respondents’ categories, except for questions regarding experience with XAI: respondents with experience using XAI methods showed a stronger preference for the ProtoPNet method, while those without such experience preferred ACE.  In general, respondents did not express a lack of trust in the ML system’s predictions, although we did not conduct additional analyses to determine whether observing the explanation increased their level of trust.

The most interesting part, however, was examining the correlations between the expected animal features previously indicated by the respondents and the best explanations they selected. Strong correlations were found between certain pairs; for example, the ACE score had the strongest correlation with the expected presence of an elephant’s trunk, while ProtoPNet was associated with the mouth in photos of hippos and the presence of stripes on tigers or zebras.

To sum up, we believe that this study indicates that human-centered evaluation should not be conducted as relatively simple user satisfaction questionnaires (often found in the current literature, see e.g., critical opinions in \cite{Lopes2022}), but rather requires careful planning and preparation of this type of research. It seems important to draw inspirations from survey research in psychology and to conduct at least a limited pilot study on a small group of people to verify the questions, the presentation format, and any potentially difficult or ambiguous elements. Furthermore, based on our experience, it is necessary to design the appropriate user interface and to provide users with the necessary information. We believe it is important to include additional questions -- as they can be useful, for example, to verify the accuracy of users’ responses.
%(e.g., we discovered few inconsistencies in the responses of participants, such as between age and education) and to confirm that their expectations align with the selected XAI methods.

\section{The Ambiguity of Counterfactual Evaluations}
\label{sec:counter}

Counterfactual explanations (briefly \textit{counterfactuals}) are another, type of post-hoc explanations for black-box model predictions~\cite{verma2020counterfactual}. Given an input instance and its corresponding prediction by the black box classifier, a counterfactual specifies the minimal changes to feature values required to obtain a different, typically more favorable, prediction from the model. Unlike other popular explanations which attempt to answer the question  ``Why this decision was made'', the counterfactuals directly address the question: ``what would need to change for the decision to be different?''. They provide actionable insight into model behavior and empower individuals to understand, and potentially contest, automated decisions in fields such as finance, healthcare, and many others \cite{counterfactuals_medical,Guidotti2022}.

Up to now, quite a high number of algorithms for generating counterfactual explanations have been introduced. They are based on different principles; for comprehensive surveys, see, e.g., \cite{Guidotti2022,verma2020counterfactual}. Depending on the specific method, some properties of counterfactuals are expected to be met.  The most common properties are \textit{validity} of the decision change, its smallest \textit{proximity} to the input instance, \textit{sparsity} of recommended changes (i.e., minimization of changed features), their \textit{actionability}, i.e., the counterfactual should not modify immutable features or violate some monotonic constraints, and \textit{plausibility} of locating the counterfactual within a high-density region of the data, ensuring that the proposed counterfactuals are realistic and feasible within the context of the observed data distribution, \textit{discriminative power} of the generated counterfactual examples should be in the region of the feature space dominated by the expected class, and others \cite{stefanowski2023,wielopolski2024probabilistically}. 

These and other properties are evaluated using appropriate evaluation measures – see the list in \cite{Guidotti2022}. It should be noted, however, that in the case of counterfactuals, there are no clear guidelines for using single evaluation measures (such as fidelity for other explanations) beyond checking validity.  Rather, depending on the chosen method of generating counterfactuals, they are designed to satisfy specific measures – this is particularly evident among optimization methods, which explicitly define specific criteria in their loss functions.  Studies introducing new methods generally demonstrate better performance on selected metrics (criteria) compared to alternative, earlier methods. Even though some of these measures are at least related to each other (e.g., such as \textit{proximity} and \textit{sparsity}), many of them are not aligned and even contradictory. %For example, \textit{proximity} encourages the generation of a counterfactual that is as close to the decision boundary as possible, whereas the \textit{discriminative power} favours explanations in dense areas dominated by the other class, which are most likely to be far from the decision boundary \cite{stefanowski2023}. 
Furthermore, experimental studies show that even very similar methods generate counterfactuals that differ significantly (due to changes in the values of specific features); see the examples in \cite{stepka2024multi}. 

Choosing a single method and a specific solution is quite a challenging task that involves finding a trade-off between conflicting aspects of explanation quality. Looking at the available experimental results, it is clear that there is no single, definitive best method \cite{Guidotti2022}.

In paper \cite{stepka2024multi}, instead of looking for single method for generating counterfactuals, we proposed to use several good methods for  providing a diversified, richer set of explanations  -- i.e., we introduce \textit{ensemble of multiple base explainers}  Then,  a multiple criteria approach is used to reduce the number of considered explanations to a smaller and concise set by using the \textit{dominance relation} and constructing a Pareto front, i.e.,~a subset of explanations that are not worse than others on at least one criterion. If their number is relatively small, then all of them could be analyzed by the user. However, when the Pareto front may contain still more solutions,  we propose to select a final counterfactual from this front by applying the \textit{multiple criteria Ideal Point Method}, which does not require additional preference elicitation from  the user, and is computationally efficient. 

Despite the simplicity of this multi-criteria approach,  the presented experiments demonstrated the utility of it over several datasets. Indeed, the selected counterfactuals were quite competitive with the solutions offered by the best of the single methods. Moreover, this kind of the proposed framework aligns with the human perspective. Humans could make trade-offs between criteria when selecting the best explanations or  decide whether they prefer only one of them.

\section{Beyond Static XAI}

The next challenge concerns the focus of current XAI methods on static data and models.
Let us note that many AI systems are increasingly deployed in non-stationary environments, where data distributions and system dynamics evolve over time. These changes arise in many real-world applications and are commonly studied under the notion of \emph{concept drift}~\cite{Read2025}, which  captures changes in the underlying data-generating process over time and its impact on the model performance.

While substantial research on evolving data has been already focused on detecting drift and adapting models to these changes~\cite{Lu2018}, significantly less attention has been paid to explaining how and why these changes occur~\cite{hinder2023model}.  

This open questions on potential usage XAI methods.  However, most these  methods are designed for static settings only, where models are trained once and explanations are generated for a fixed data distribution and model. When applied in evolving systems they could provide outdated information or lead to  incorrect interpretations

Existing attempts to extend XAI to evolving settings include snapshot-based analyses that repeatedly apply static explanation methods at different time points, or are preliminary incremental attempts (see their survey in \cite{Pelosi2025}). They are still limited for explicitly explaining how and why system behaviour changes over time and what are the cause of it with respect to changes in underlying data distribution in evolving data streams.

In the work \cite{karolczak2025explaining}, we considered stream adaptation of prototypes explanations representing predictions of tree ensembles such as bagging or random forests. In particular we studied different aspects of their evaluation with respect to changes in the number of generated prototypes, and their positions in feature space or content of their neighborhood. Consequently, we introduced a set of measures for comparing two sets of prototypes, possibly before and after a change in the data stream. Some of the proposals were inspired by cluster quality measures. However, more relevant were new proposals such as mean minimal distance between prototypes, mean prototype centroid displacement, and prototype reassignment impact. Experiments on synthetic and real-world data streams demonstrated that they could support detecting concept drift. Moreover, we showed that combining these measures  with feature parallel plots can support the analysis of the causes of drift concerning the most influential features, which opens avenues for more advanced methods to identify drift type and location.
  
Yet another attempt was undertaken in our recent paper on  explaining concept drift through the analysis of evolution of group counterfactuals \cite{stkepka2026explaining}. Although most research concerns individual CEs \cite{Guidotti2022}, a quite newer line of work focuses on group counterfactual explanations (GCEs), which extend per-sample CEs to groups of similar instances sharing a common explanation vector, see a review in \cite{furman2024unifying}. In our study we generalize GLANCE algorithm to generate GCEs \cite{kavouras2026glance},  then in consecutive time steps we  studied the evolution of cluster centroids and their associated counterfactual vectors, in particular before and after a drift, showing that they could serve as an interpretable proxy for assessing locality of drifts and examine shifts in cluster positions to check if the decision boundary has moved. Significant differences in vectors suggest a larger shift. We quantify this difference by calculating the cosine similarity between vectors and visualizing it per-group. Additionally, we could visualize the feature-wise vector directions to identify which features were most affected. 

We also examined changes occurring at the data level itself (by using various descriptive statistical tools to examine changing data distributions) and model performance (here, through the simplest analysis of decreases in model predictions—which can be achieved using drift detectors).  As we showed in \cite{stkepka2026explaining}  although these layers of information correspond to different manifestations of changes their interaction and the joint analysis could provide a more comprehensive understanding the drift that analyzing each part independently.

\section{Open Issues and Final Remarks}

We discussed several issues of the evaluation of explainable AI -- both in the context of static and evolving data. Firstly we should stress that a well defined, unified, “machine-learning-style” evaluation framework for XAI is still missing. We should  conduct more fundamental research on new experimental standards. There are several open issues such as:
\begin{itemize}
\item Current evaluation measures depend very much on the type of explanation and the tasks - they do not generalize.
\item Most of them are a kind functional proxies for local post-hoc explainers and as they only approximate fairly complex and hidden behaviors of the models - more comparative studies should be undertaken.
\item Knowing their limitations, the wider and more diversified set of measures can be applied to evaluate XAI methods also with inspirations from the multi-criteria decision analysis.
\item The most current research covers mainly local post-hoc explanations, while the global ones are less studied.
\item There is lack of enough benchmark datasets with annotated information on the ground-truth for explanations; More work on them, also with synthetic generators, are necessary.
\end{itemize}

Explanations must serve humans. Although application- and human- grounded evaluations are the most important, they are too neglected in current XAI research. More systemic work on human studies is needed  -- also taking inspirations from the experiences of experimental psychology and existing frameworks or protocols for designing and conducting survey research. In general,  more interdisciplinary research should be carried in XAI field. 

Other research questions on human grounded studies could cover: How to organize interaction sessions with users? How should we prepare users for them? How should we validate the correctness of their answers. Whether we should move into NLP sequential dialog, instead of simple one-step answers? How could we better incorporate cause-effect relationships (causality) and background knowledge? And a more general research question - Do the explanations really increase human’s trust in predictions and using AI systems?

Moreover, we pointed out the difficult challenges of explaining model performance in dynamic environments (such as evolving data streams with concept drifts or continual learning).  Evaluating such changes, also for changing explanations over time, remains one of the most challenging aspects for XAI. Existing evaluation measure are focused on static setting and measure basic properties at single points in time, while we need to take into account issues of longer temporal dynamics (and also possible interactions between data, model and explanation changes). This area essentially requires more intensive fundamental research and interesting application problems to verify new proposed ideas.

We hope that air space exploration, earth observation, and various sub-fields of medicine - as considered in the EASi-EXPLIMED IJCAI-ECAI workshop - may provide inspirations to the aforementioned research challenges.
 
\smallskip

\noindent
\textbf{Acknowledgment} This work was fully funded by the National Science Centre, Poland, under the OPUS programme (grant number 2023/51/B/ST6/00545).

\bibliographystyle{splncs04}
\bibliography{myijcai}

\end{document}